# Jointly Learning Structured Analysis Discriminative Dictionary and Analysis Multiclass Classifier

Zhao Zhang, *Member*, *IEEE*, Weiming Jiang , Jie Qin, Li Zhang, *Member*, *IEEE*, Fanzhang Li , Min Zhang, and Shuicheng Yan, *Fellow*, *IEEE*

*Abstract*— In this paper, we propose an analysis mechanism based structured *Analysis Discriminative Dictionary Learning* (ADDL) framework. ADDL seamlessly integrates the analysis discriminative dictionary learning, analysis representation and analysis classifier training into a unified model. The applied analysis mechanism can make sure that the learnt dictionaries, representations and linear classifiers over different classes are independent and discriminating as much as possible. The dictionary is obtained by minimizing a reconstruction error and an analytical incoherence promoting term that encourages the sub-dictionaries associated with different classes to be independent. To obtain the representation coefficients, ADDL imposes a sparse $l_{2,1}$-norm constraint on the coding coefficients instead of using $l_0$ or $l_1$-norm, since the $l_0$ or $l_1$-norm constraint applied in most existing DL criteria makes the training phase time consuming. The codes-extraction projection that bridges data with the sparse codes by extracting special features from the given samples is calculated via minimizing a sparse codes approximation term. Then we compute a linear classifier based on the approximated sparse codes by an analysis mechanism to simultaneously consider the classification and representation powers. Thus, the classification approach of our model is very efficient, because it can avoid the extra time-consuming sparse reconstruction process with trained dictionary for each new test data as most existing DL algorithms. Simulations on real image databases demonstrate that our ADDL model can obtain superior performance over other state-of-the-arts.

*Index Terms*— Structured analysis discriminative dictionary learning, projective sparse representation, analysis multiclass classifier, analytical incoherence promotion[1]

## I. INTRODUCTION

Sparse representation (SR) has achieved a great success in various practical applications in communities of computer vision and learning systems [1-8][49-53]. SR reconstructs an input data by a linear combination of a few items from a dictionary [10-12], so the dictionary is crucial to the process of SR. Wright et al. [9] utilizes the entire set of training data as a dictionary for representing face images and delivers an impressive face recognition result, but obtaining the sparse representation coefficients from the dictionary (i.e., training set) may suffer from several drawbacks, i.e., most real data are usually noisy or corrupted [1], which may decrease the performance; the dictionary of large size makes the process of obtaining the process of SR inefficient [1-2], which may constrain its real applications. To tackle this issue, several efforts on the study of compact or over-complete dictionary learning (DL) have been made in recent years, for instance [1-3], [7-8], [10-15], etc.

The existing DL criteria designed to compute dictionaries that are suitable for representation and classification can be generally divided into two categories, i.e., unsupervised and discriminant ones. Unsupervised methods do not use any supervised prior information (e.g., class label information) about training data and aim to minimize the residual error of reconstructing the given data to produce a dictionary, e.g., [13][16-19], among which *K-Singular Value Decomposition* (KSVD) [13] is one most representative DL method. KSVD mainly aim to learn an over-complete dictionary from the training data via generalizing the *k*-means clustering. But note that KSVD hopes the learned dictionary can represent the training data effectively, i.e., it is not suitable for dealing with the classification task. In contrast, by utilizing the label information of training data, discriminative DL methods are proved to be more effective to enhance classification [2][10][14][20]. Concretely speaking, existing supervised methods can be divided into three categories [2]. The first type of DL methods selects the dictionary items from an initially large dictionary to produce a compact dictionary [21-22], which suffer from some obvious drawbacks, e.g., computationally expensive due to the dictionary of large-size is required to ensure the discriminating capability [2]. The second kind incorporates discriminative terms into the objective function, such as [1-2], [7-8], [10-12], [14], [23-24], among which *Discriminative KSVD* (D-KSVD) [14], *Label Consistent KSVD* (LC-KSVD) [2] and *Fisher Discrimination Dictionary Learning* (FDDL) [1] are three representative ones. D-KSVD includes the classification error into KSVD to enhance classification, LC-KSVD incorporates a label consistent constraint further into KSVD to guarantee the discrimination of learned representation, while FDDL seeks a structured dictionary and force the coding coefficients to deliver small within-class scatter and large between-class scatter. The third kind aims at computing category-specific dictionaries to encourage each sub-dictionary to correspond to a single class, for instance [3], [15], [25-26]. In [15], an incoherence promoting term is incorporated to ensure the learned class-specific dictionaries are independent. Zhou et al. [25] have proposed a DL algorithm by learning multiple dictionaries for correlated objective categories. Gu et al. [3] have also extended the traditional discriminative synthesis dictionary learning to discriminative synthesis and analysis dictionary pair learning for representation.

It is worth noting that most aforementioned DL methods still suffer from some drawbacks. First, to obtain the sparse codes, most existing methods adopt $l_0$ or $l_1$-norm sparsity constraint on the coefficients, resulting in time-consuming training phase. Second, to deal with the classification task, existing DL methods usually contain two separable phases:

———————————
Z. Zhang, W. Jiang, L. Zhang, F. Li and M. Zhang are with the School of Computer Science and Technology & Provincial Key Laboratory for Computer Information Processing Technology, Soochow University, Suzhou 215006, China (e-mails: cszzhang@gmail.com, cswmjiang@gmail.com, {zhangliml, lfzh, minzhang}@suda.eduu.cn)
J. Qin is with the Computer Vision Laboratory, ETH Zürich, 8092 Zürich, Switzerland (e-mail: jqin@vision.ee.ethz.ch).
S. Yan is with the Department of Electrical and Computer Engineering, National University of Singapore, Singapore. (e-mail: eleyans@nus.edu.sg)

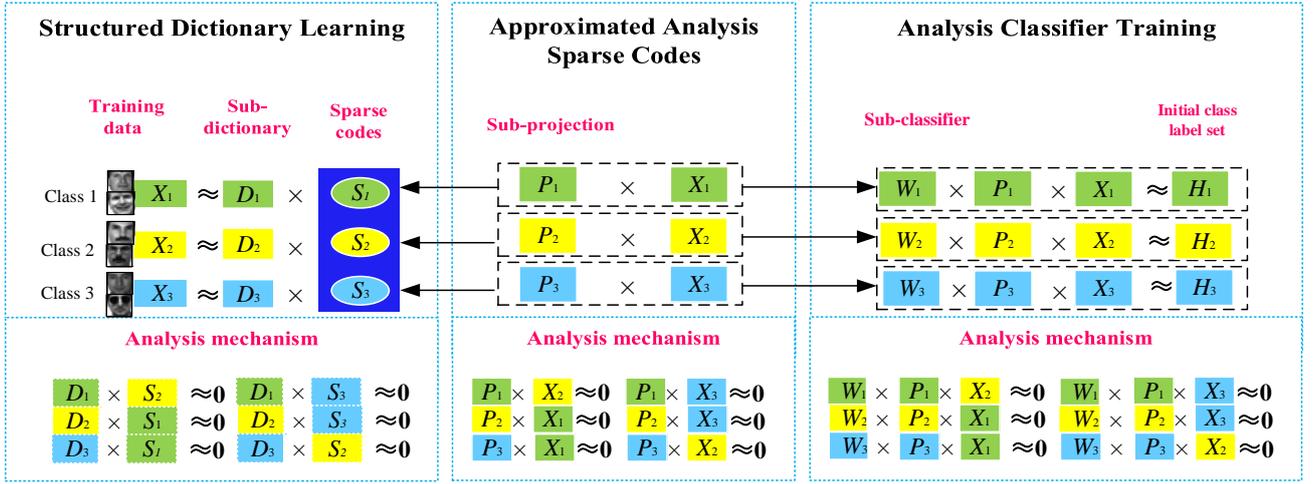

**Fig. 1**: Overview of our proposed analysis mechanism based discriminative structured DL framework.

representation learning (i.e., coding) and classifier training. Specifically, the query data is firstly encoded over the learnt dictionary with the sparsity constraint, then classification is performed over the coding coefficients, which is clearly not straightforward and inefficient. *Projective Dictionary Pair Learning* (DPL) [3] seeks a synthesis dictionary for coding and an analysis dictionary to learn the coding coefficients without imposing the costly $l_0/l_1$-norm constraint, which has greatly reduced the time complexity in training and testing, but it still has some shortcomings. First, DPL cannot ensure the learnt codes to satisfy the sparse property that can lead to natural discriminating power and accurate predictions [1]; Second, DPL does not consider the relationship between the synthesis sub-dictionaries of various classes; Third, DPL aims to classify each new data by minimizing the residual between a test signal and its approximation by the synthesis sub-dictionary and analysis sub-dictionary associated with each class, which is also not straightforward.

In this paper, we thus propose a new analysis mechanism based structured discriminative DL framework to enhance classification and reduce the computational cost at the same time. In the following, we present the major contributions. (1) Technically, a joint *Analysis Discriminative Dictionary Learning* (ADDL) and projective sparse representation model is derived. ADDL seamlessly integrates the analysis discriminative dictionary learning, analysis representation and analysis classifier training into a unified model. ADDL learns each sub-dictionary separately for reconstructing data within a class, and also learns independent sub-projections to extract the representation coefficients of intra-class data, similarly as DPL [3]. To compute the sparse codes without using the costly $l_0/l_1$-norm constraint, a $l_{2,1}$-norm constraint is used on the coefficients, where the $l_{2,1}$-norm constraint can ensure the coefficients to be sparse in rows [27-28]. Due to the joint learning of analysis classifier, our method avoids involving the extra time-consuming reconstruction process with trained dictionary for each new data, suffered in most existing discriminative DL methods [2][9-10][14]. (2) We incorporate an analytical incoherence promoting term into the criterion to minimize the reconstruction error over the sub-dictionary associated with each class *i* and the sparse codes of training signals that do not belong to class *i*. As a result, class-specific dictionaries can be learnt from the optimizations and the sub-dictionaries are independent as much as possible. We also design a new analytical classifier training term to enhance the discriminating power of learnt classifier. (3) Based on the analysis mechanism, our ADDL can compute a structured discriminative dictionary, a set of projective sparse coding coefficients, and an analysis linear classifier jointly with low cost from one objective function. More specifically, the used analysis mechanism can ensure the learnt dictionaries, representations and classifiers are independent and discriminating as much as possible.

To clearly demonstrate our proposed analysis mechanism, we illustrate the overview of our approach in Fig. 1, where we consider a three-class case as an example, $X_l$ is the training set of class $l$, $D_l, S_l, P_l, W_l$ and $H_l$ represent the sub-dictionary, sparse coding coefficients, sub-projection, sub-classifier and initial class label set within each class $l$, respectively. As can be seen clearly, $P_l$ can bridge the data with sparse codes, i.e., the approximated sparse codes $P_l X_l$ can be obtained by embedding given data onto $P_l$, and the approximated $P_l X_l$ are then applied to train the analysis discriminative classifier *W*. Note that the presented analysis mechanism can be regarded as a discriminative mechanism, i.e., we hope that the reconstruction $D_l S_j$, the reconstruction $P_l X_j$, and the reconstruction $W_l P_l X_j$ where $l \neq j$, are all nearly null sparse, which can ensure each sub-dictionary, sub-projection and sub-classifier within each class $l$ to be independent as much as possible.

The paper is outlined as follows. Section II reviews the related works. In Sections III, we propose ADDL. Section IV shows the connections with other DL methods. Section V describes the settings and evaluates our methods. Finally, the paper is concluded in Section VI.

## II. RELATED WORK

### A. Dictionary Learning for Representation

Let $X = [X_1, \cdots X_l, \cdots, X_c] \in \mathbb{R}^{n \times N}$ be a set of training samples from *c* classes, where $X_l \in \mathbb{R}^{n \times N_l}$ is the training set of class *l*, $N_l$ is the number of data in class *l*, i.e., $\sum_{l=1}^{c} N_l = N$, and *n* is the original dimensionality. By computing a reconstructive dictionary with *K* items to gain the sparse representation of *X*, most existing dictionary learning algorithms solves the following general optimization problem:

$$\langle D, S \rangle = \arg\min_{D,S} \|X - DS\|_F^2 + \lambda \|S\|_p, \quad (1)$$

where $\lambda > 0$ denotes a scalar constant, $D = [d_1, \cdots d_K] \in \mathbb{R}^{n \times K}$ is a synthesis dictionary, which is obtained by minimizing the reconstruction error $\|X - DS\|_F^2$. $\|S\|_p$ denotes the $l_p$-norm regularized term, where $S = [s_1 \cdots s_N] \in \mathbb{R}^{K \times N}$ are the coding coefficients of $X$ and the parameter $p$ is usually set to $p = 0$ or 1 to ensure the learnt coefficients to be sparse, but the computation is usually costly. To solve the inefficient issue, Gu. et al. [3] proposed to learn a synthesis dictionary $D$ and an analysis dictionary $P$ without using the costly $l_0/l_1$-norm constraint by solving the following problem:

$$\langle P, D \rangle = \arg\min_{P,D} \sum_{l=1}^{c} \|X_l - D_l P_l X_l\|_F^2 + \lambda \|P_l \overline{X}_l\|_F^2,$$
$$s.t. \|d_i\|_2^2 \leq 1 \quad (2)$$

where $\overline{X}_l$ is the complementary data matrix of $X_l$ in $X$, i.e., excluding $X_l$ itself from $X$. $D_l \in \mathbb{R}^{n \times k}$ is synthesis dictionary of class $l$ with $k$ items, $P_l \in \mathbb{R}^{k \times n}$ is an analysis dictionary of class $l$. So, $D = [D_1, \cdots D_l, \cdots D_c]$, $P = [P_1; \cdots P_l; \cdots P_c] \in \mathbb{R}^{K \times n}$. $d_i$ is the $i$-th atom of dictionary $D$, the constraint $\|d_i\|_2^2 \leq 1$ is to avoid the trivial solution $P_l = 0$ and make the computation stable. This formulation is based on (1), but improves (1) by finding an analysis dictionary $P$ to analytically obtain the coding coefficients by $S = PX$. More specifically, by setting $P_l X_j \approx 0, \forall j \neq l$, the resulted approximate coefficient matrix $PX$ will be nearly block diagonal.

### B. Dictionary Learning for Classification

Given a new test data $x_{new}$, its category can be identified by using two widely used approaches in DL. The first kind is SRC-like method [9]. That is, the coefficient $s_{new}$ of $x_{new}$ is firstly computed over the learnt dictionary. Then $x_{new}$ can be classified by minimizing the following residual:

$$identity(x_{new}) = \arg\min_l \|x_{new} - D\delta_l(s_{new})\|_2, \quad (3)$$

where $\delta_l(s_{new})$ is a vector whose nonzero entries in $s_{new}$ are associated with class $l \in \{1, \cdots c\}$. That is, $x_{new}$ is classified based on the approximated coding coefficients (i.e., sparse codes) by assigning the codes to the object class that can minimize the residual [9]. Note that the algorithms in this kind do not construct an explicit classifier and is essentially a lazy classifier [31]. The other kind is to jointly compute a dictionary $D$ and a linear classifier $W \in \mathbb{R}^{c \times K}$, e.g., [2][10][14][29-30]. In this scenario, a unified problem for learning $D$ and $W$ jointly can be formulated as

$$\langle D, S, W \rangle = \arg\min_{D,S,W} \|X - DS\|_F^2 + \lambda \|S\|_p + \sum_i \Psi\{h_i, f(s_i, W)\}, \quad (4)$$

where $\Psi$ indicates the classification loss function, $h_i$ is the pre-defined label of data $x_i$. Thus, $x_{new}$ can be classified by embedding its coefficient $s_{new}$ to the classifier. It is worth noting that the above two methods classify each test signal based on the coding coefficient of this signal, i.e., an extra time consuming sparse reconstruction process for each test data is involved for classification,. Although DPL learns an analysis dictionary that can predict the coding coefficients simply, it classifies each new test signal by minimizing the residual between the test signal and its approximation by using the synthesis sub-dictionary that is associated with each class, which is not straightforward as well.

## III. STRUCTURED ANALYSIS DISCRIMINATIVE DICTIONARY LEARNING (ADDL)

### A. The Objective Function

In this paper, we mainly propose a new analysis mechanism based structured discriminative DL model for representation and classification by jointly learning a structured dictionary, a set of projective sparse coding coefficients and an analysis linear multi-class classifier. Different from the existing DL approaches that adopt the costly $l_0$ or $l_1$-norm constraint for sparse coding, our method adopt a $l_{2,1}$-norm regularizer to constrain the representation coefficients, since the $l_{2,1}$-norm constraint can ensure the coefficients to be sparse in rows [27-28], and more importantly the optimization of solving $l_{2,1}$-norm is efficient. In addition, an analysis incoherence promoting term, an analysis sparse codes extraction term, and an analysis classifier training term are incorporated to enhance the discriminating power of the learned dictionary, sparse codes and classifier at the same time. Thus, we can define the following optimization problem for ADDL:

$$\langle D, S, P, W \rangle = \arg\min_{D,S,P,W}$$
$$\sum_{l=1}^{c} \left\{ \|X_l - D_l S_l\|_F^2 + \alpha f(D_l) + \tau r(P_l, S_l) + \lambda g(H_l, W_l, P_l) \right\}, \quad (5)$$
$$s.t. \|d_v\|_2^2 \leq 1, v \in \{1, \cdots K\}$$

where $f(D_l)$ is the analysis incoherence promoting function, $r(P_l, S_l)$ is the analysis sparse codes extraction function, and $g(H_l, W_l, P_l)$ is the analysis classifier training function. $\alpha, \tau$ and $\lambda$ are three positive scalar parameters associated with the above three function terms. Next, we will describe the formulations of $f(D_l)$, $r(P_l, S_l)$ and $g(H_l, W_l, P_l)$.

#### 1) Analysis incoherence promoting function

Suppose $D = [D_1, \cdots D_l, \cdots D_c] \in \mathbb{R}^{n \times K}$ is the learned dictionary, where $D_l \in \mathbb{R}^{n \times k}$ is the sub-dictionary corresponding to the class $l$, $S_l \in \mathbb{R}^{k \times N_l}$ is the coding coefficient of $X_l$ over sub-dictionary $D_l$. Note that the coefficient $S_l$ should be able to well represent $X_l$, i.e., $X_l \approx D_l S_l$. More importantly, $S_l$ is corresponded to class $l$, so it is expected that $X_l$ can be well represented by $S_l$ but not by $S_j, j \neq l$, thus we want $\|D_l S_j\|_F^2$ to be small during the optimization to ensure that $D_l S_j$ is not close to $X_l$, which enhances the discriminating power of learned dictionary. So, we define the following analysis incoherence promoting function for our ADDL method:

$$f(D_l) = \|D_l \overline{S}_l\|_F^2, \quad (6)$$

where $\overline{S}_l$ represents the complementary matrix of $S_l$ in the whole coding coefficient matrix $S = [S_1, \cdots, S_c]$. The term $\|D_l \overline{S}_l\|_F^2$ is the analysis incoherence promoting term.

#### 2) Analysis sparse codes extraction function

Denote by $P = [P_1; \cdots P_l; \cdots P_c] \in \mathbb{R}^{K \times n}$ the underlying projection to extract the representation coefficients, where $P_l \in \mathbb{R}^{k \times n}$ is the sub-projection corresponding to class $l$. In this case, we hope that the sub-projection $P_l$ can bridge signals with the approximated coding coefficients by calculating the special features from given data [1]. That is, we hope that the term $P_l X_l$ can well approximate the coefficients $S_l$:

$$P_l X_l \approx S_l. \quad (7)$$

Note that by minimizing this approximation error we can obtain an optimal sub-projection $P_l$ that can bridge data with the sparse codes within class $l$, but note that we cannot ensure $P_l$ will not bridge data of class $j$, where $j \neq l$. To solve this issue, inspired by [3] we hope $P_l$ can project the data from class $j$, $j \neq l$, to a nearly null space, that is,

$$P_l X_j \approx 0, \forall j \neq l. \quad (8)$$

At the same time, the coefficients $S_l$ should be sparse as much as possible, so we can also term $S_l$ as sparse codes.

Most existing models use the $l_0$ or $l_1$-norm regularization on the coefficients, which usually results in time consuming computations. Note that $l_{2,1}$-norm regularization can ensure the coefficients to be sparse in rows as well [27-28] and the computation of solving the $l_{2,1}$-norm is usually efficient and easy, thus we use the $l_{2,1}$-norm constraint on $S_l$ to ensure the sparsity property. Finally, we can define the following analysis sparse codes extraction function:

$$r(P_l, S_l) = \|P_l X_l - S_l\|_F^2 + \|P_l \overline{X_l}\|_F^2 + \|S_l\|_{2,1}, \quad (9)$$

where $\|S_l\|_{2,1}$ is $l_{2,1}$-norm regularization on the coefficients, and $\|P_l \overline{X_l}\|_F$ is the analysis sparse codes extraction term.

*3) Analysis multiclass classifier training function*

Suppose that $X_l = [x_{l,1}, \cdots, x_{l,i}, \cdots, x_{l,N_l}]$, where $x_{l,i} \in \mathbb{R}^n$ is an input data. We introduce a variable label vector for each $x_{l,i}$ as $h_{l,i} = [0, \cdots 1, \cdots 0]^T \in \mathbb{R}^c$, where the position of nonzero value indicates the class assignment of $x_{l,i}$. We also assume that its label vector can be approximated by its embedded coefficient $P_l x_{l,i}$ via a linear sub-classifier $W_l \in \mathbb{R}^{c \times k}$, i.e., $h_{l,i} \approx W_l P_l x_{l,i}$. Suppose $H_l = [h_{l,1}, \cdots h_{l,N_l}] \in \mathbb{R}^{c \times N_l}$ are the class labels of $X_l$, the approximation can be formulated as

$$H_l \approx W_l P_l X_l. \quad (10)$$

In order to enhance the discriminating power of learnt sub-classifiers $W_l$, we similarly hope that $W_l$ can predict the labels of samples from class $j$ to a nearly null space:

$$W_l P_l X_j \approx 0, \forall j \neq l. \quad (11)$$

Therefore, the analysis classifier training function can be defined as follows:

$$g(H_l, W_l, P_l) = \|H_l - W_l P_l X_l\|_F^2 + \|W_l P_l \overline{X_l}\|_F^2, \quad (12)$$

where $\|W_l P_l \overline{X_l}\|_F^2$ is the analysis classifier training term. By combing (5), (6), (9) and (12), the final objective function of our ADDL can be reformulated from (5) as

$$\langle D, S, P, W \rangle = \arg\min_{D,S,P,W} \sum_{l=1}^{c} \Big\{ \|X_l - D_l S_l\|_F^2 + \alpha \|D_l \overline{S_l}\|_F^2 \\
+ \tau \Big[ \|P_l X_l - S_l\|_F^2 + \|P_l \overline{X_l}\|_F^2 + \|S_l\|_{2,1} \Big] \\
+ \lambda \Big[ \|H_l - W_l P_l X_l\|_F^2 + \|W_l P_l \overline{X_l}\|_F^2 \Big] \Big\}, \quad (13)$$

$$s.t. \|d_v\|_2^2 \leq 1, v \in \{1, \ldots, K\}$$

where $\|d_v\|_2^2 \leq 1$ is a dictionary atom constraint for making the computation of ADDL stable [3][48]. To illustrate the effect of this constraint, we use the ORL face dataset [32] as an example to compute the value of $\|d_v\|_2^2$, $v \in \{1, \ldots, K\}$. This dataset contains 40 subjects with 10 images per subject. All the images were resized to 32x32. We randomly choose half of images per subject for training and the other half of data for testing. The values are illustrated in Fig. 2, from which we can see all the values are smaller than 1. Since we have incorporated an incoherence term to promote the learned sub-dictionary as independent as possible, the values of $\|d_v\|_2^2$ are not fixed. Besides, we use an analysis mechanism to enhance the discriminating power of learnt dictionary $D$, sparse codes extraction $P$, and classifier $W$. To illustrate the superiority of our model by using the analysis mechanism, we also use the ORL face dataset as an example to visualize the sparse codes and the computed soft labels. Fig.3 (a) illustrates the approximated coding coefficients ($PX$) of the testing data. We can find that the approximated coding coefficient matrix of our ADDL algorithm is nearly block-diagonal, and the coding coefficient of each image is sparse. In contrast, if we do not use the analysis mechanism, the coding coefficients cannot be well structured, as can be seen from Fig. 3(b). Fig. 4(a) illustrates the computed soft labels of testing data. We can observe that the testing data can be well induced by our ADDL. In contrast, as shown in Fig. 4(b), if we do not employ the analysis mechanism, the resulted classifier would have a weak discriminating power. Clearly, based on using the proposed analysis mechanism, the computed coding coefficients and soft labels are more accurate, which can directly enhance the performance.

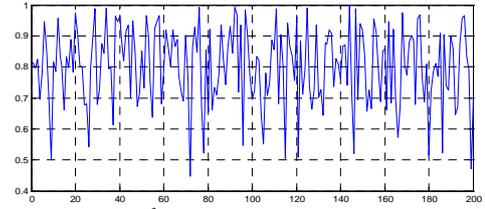

**Fig. 2:** The values of $\|d_v\|_2^2$, where the *x*-axis denotes value of $v$, *y*-axis denotes the value of $\|d_v\|_2^2$. Parameters $\alpha=0.1$, $\tau=0.1$ and $\lambda=0.1$.

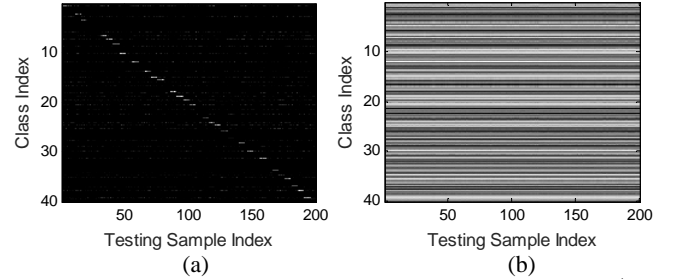

**Fig. 3:** (a) The approximated sparse representation coefficients ($P^*X$) by our ADDL with analysis mechanism; (b) The approximated sparse coefficients without the analysis mechanism. The parameters $\alpha=0.1$, $\tau=0.1$ and $\lambda=0.1$.

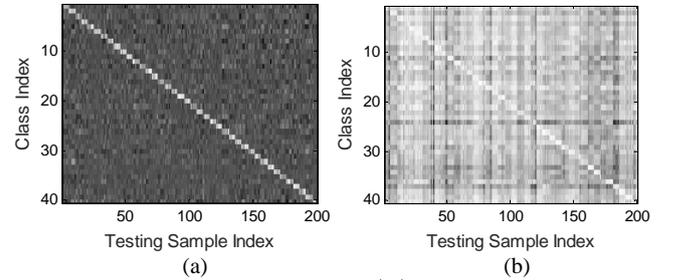

**Fig. 4**: (a) The computed soft labels ($W^*P^*X$) by ADDL with analysis mechanism; (b) The gained soft labels without analysis mechanism. The parameters $\alpha=0.1, \tau=0.1$ and $\lambda=0.1$.

*B. Optimization*

We show the optimization of our ADDL. We first initialize $D$, $P$, and $W$ as random matrices with unit F-norm, then the minimization can be alternated between the following steps:

**1) Fix $D$ and $P$, optimize $S$ and $\Lambda$:** According to the definition of $l_{2,1}$-norm [27-28], i.e., $\|S\|_{2,1} = 2tr(S^T \Lambda S)$, where $\Lambda$ is a diagonal matrix with entries $\Lambda_{ii} = 1/2\|S^i\|_2$ where $S^i$ is the *i*-th row of $S$. By removing terms that are irrelevant to $S$, the problem of our ADDL can be simplified as

$$S^* = \arg\min_S \wp(S), \text{ where}$$

$$\wp(S) = \sum_{l=1}^{c} \Big\{ \|X_l - D_l S_l\|_F^2 + \tau \|P_l X_l - S_l\|_F^2 + \tau \|S_l\|_{2,1} \Big\} \quad (14)$$

$$= \sum_{l=1}^{c} \Big\{ \|X_l - D_l S_l\|_F^2 + \tau \|P_l X_l - S_l\|_F^2 + \tau tr(S_l^T \Lambda_l S_l) \Big\}$$

As each $\Lambda^i \neq 0$, by setting the derivative $\partial \wp(S)/\partial S = 0$, we can have the following closed-form solution:

$$S_l^* = \left(D_l^\mathrm{T} D_l + \tau I + \tau \Lambda_l\right)^{-1}\left(\tau P_l X_l + D_l^\mathrm{T} X_l\right). \quad (15)$$

Then, we can update the matrix $\Lambda$ by using $\Lambda_{ii} = 1/2\|S^i\|_2$.

**2) Fix $S$ and $W$, optimize $P$:** By removing terms that are irrelevant to $P$, we have the following reduced problem:

$$P^* = \arg\min_P \Im(P), \text{ where } \Im(P) = $$
$$\sum_{l=1}^c \left\{\tau\|P_l X_l - S_l\|_F^2 + \tau\|P_l \overline{X_l}\|_F^2 + \lambda\|H_l - W_l P_l X_l\|_F^2 + \lambda\|W_l P_l \overline{X_l}\|_F^2\right\}$$
(16)

By setting the derivative $\partial \Im(P)/\partial P = 0$, the closed-form solution for $P$ can be obtained as

$$P_l^* = \left(\tau I + \lambda W_l^\mathrm{T} W_l\right)^{-1}\left(\tau S_l X_l^\mathrm{T} + \lambda W_l^\mathrm{T} H_l X_l^\mathrm{T}\right)$$
$$\times \left(X_l X_l^\mathrm{T} + \overline{X_l}\,\overline{X_l}^\mathrm{T} + \gamma I\right)^{-1}, \quad (17)$$

where $\gamma = 1e-4$ is a small number. Due to the fact that the dimension of feature space may be higher than the number of samples, the inverse of $XX^\mathrm{T}$ may be singular, so $\gamma I$ is added to avoid the singularity issue in real problems [3].

**3) Fix $P$, optimize $W$:** By removing the terms that are irrelevant to $W$, we have the following formulation:

$$W^* = \arg\min_W \Phi(W)$$
$$\text{where } \Phi(W) = \sum_{l=1}^c \left\{\|H_l - W_l P_l X_l\|_F^2 + \|W_l P_l \overline{X_l}\|_F^2\right\}, \quad (18)$$

By setting $\partial \Phi(W)/\partial W = 0$, we can update $W$ by using

$$W_l^* = H_l X_l^\mathrm{T} P_l^\mathrm{T}\left(P_l^\mathrm{T} X_l X_l^\mathrm{T} P_l^\mathrm{T} + P_l^\mathrm{T} \overline{X_l}\,\overline{X_l}^\mathrm{T} P_l^\mathrm{T} + \gamma I\right)^{-1}, \quad (19)$$

where $\gamma I$ can also make the inverse computation stable.

**4) Fix $S$, optimize $D$:** The problem of updating $D$ can be formulated as the following problem:

$$D^* = \arg\min_D \Delta(D) = \sum_{l=1}^c \|X_l - D_l S_l\|_F^2 + \alpha\|D_l \overline{S_l}\|_F^2$$
$$\text{s.t. } \|d_v\|_2^2 \leq 1, v \in \{1,\cdots K\} \quad (20)$$

Note that the Lagrange dual function can be used to solve (20), then we can have

$$g(\eta) = \inf\left\{\|X_l - D_l S_l\|_F^2 + \alpha\|D_l \overline{S_l}\|_F^2 + \sum_{i=1}^k \eta_{l,i}\left(\|d_i\|_2^2 - 1\right)\right\}, \quad (21)$$

where $\eta_{l,i}$ denotes the Lagrange multiplier of $i$-th equality constraint $\left(\|d_i\|_2^2 - 1 \leq 0\right)$. If we construct a diagonal matrix $\mathrm{M}_l \in \mathbb{R}^{k \times k}$ with the diagonal entry being $(\mathrm{M}_l)_{ii} = \eta_i$, then (21) can be reformulated as follows:

$$L(D_l,\eta) = \|X_l - D_l S_l\|_F^2 + \alpha\|D_l \overline{S_l}\|_F^2 + tr(D_l^\mathrm{T} D_l \mathrm{M}_l) - tr(\mathrm{M}_l). \quad (22)$$

Based on setting the derivative $\partial L(D_l,\eta)/\partial D_l = 0$, we can obtain the following closed-form solution for $D_l$:

$$D_l^* = X_l S_l^\mathrm{T}\left(S_l S_l^\mathrm{T} + \alpha \overline{S_l}\,\overline{S_l}^\mathrm{T} + \mathrm{M}_l\right)^{-1}, \quad (23)$$

For the sake of reducing the computational complexity, we mainly follow [10] to discard $\mathrm{M}_l$. But note that $S_l S_l^\mathrm{T} + \alpha \overline{S_l}\,\overline{S_l}^\mathrm{T}$ cannot be ensured to be invertible and the inverse operation on $S_l S_l^\mathrm{T} + \alpha \overline{S_l}\,\overline{S_l}^\mathrm{T}$ still may produce the singular issue even if $S_l S_l^\mathrm{T} + \alpha \overline{S_l}\,\overline{S_l}^\mathrm{T}$ is invertible, so we add a regularization term $\gamma I$ into $S_l S_l^\mathrm{T} + \alpha \overline{S_l}\,\overline{S_l}^\mathrm{T}$ for avoiding the singular issue and making the computation stable, similarly as [3].

For complete presentation of the method, we summarize the optimization procedures of our ADDL in Table I, where the iteration stops as the difference between consecutive $P$ in adjacent iterations is less than 0.001 in simulations.

**Table I**: Analysis Discriminative Dictionary Learning

**Input:** Training data matrix $X$, initial label matrix $H$, parameters $\alpha$, $\tau$, $\lambda$ and dictionary size $K$.
**Output:** $D, S, P, W$.
1: Initialize $D^{(0)}$, $P^{(0)}$ and $W^{(0)}$ as the random matrices with unit F-norm; initialize $\Lambda = I$; $t=0$;
2: **while** not converge **do**
3: $\quad t \leftarrow t+1$;
4: $\quad$ Update the sparse codes matrix $S^{(t)}$ by (15);
5: $\quad$ Update the matrix $\Lambda^{(t)}$ by $\Lambda_{ii} = 1/2\|S^i\|_2$;
6: $\quad$ Update the projection matrix $P^{(t)}$ by (17);
7: $\quad$ Update the linear classifier $W^{(t)}$ by (19);
8: $\quad$ Update the dictionary $D^{(t)}$ by (23);
9: **end while**

### C. Convergence Analysis

Our proposed ADDL is an alternate convex search (ACS) algorithm [33-35]. Thus, we have the following remarks [33-35] for proving the convergence of our ADDL.

*Theorem 1* [35]. If $B \subseteq \mathbb{R}^{n \times m}, f: B \to \mathbb{R}$ is bounded and the optimizations of variables in each iteration are solvable, the generated sequence $\{f(z_i)\}_{i \in t}$ ($z_i \in B$) by ACS algorithm will converge monotonically.

*Theorem 2* [35]. Let $X \subseteq \mathbb{R}^n$, $Y \subseteq \mathbb{R}^m$ be closed sets and let $f: X \times Y \to \mathbb{R}$ be continuous. Let the optimization of each variable in each iteration be solvable. Then we have: (1) If the sequence $\{z_i\}_{i \in t}$ by ACS is contained within a compact set, the sequence will contain at least one accumulation point; (2) For each accumulation point $z^*$ of sequence $\{z_i\}_{i \in t}$: (a) if the optimal solution of one variable with others fixed at each iteration is unique, then all accumulation points will be the local optimal solutions and have the same function value; (b) if the optimal solution of each variable is unique, then we have $\lim_{i \to \infty}\|z_{i+1} - z_i\| = 0$, and the accumulation points can form a compact continuum $C$.

Next, we can present some remarks on the convergence of our ADDL and analyze the property of our ADDL.

*Remark 1.* The generated sequence $\{f(D^i, S^i, \Lambda^i, P^i, W^i)\}_{i \in t}$ by our ADDL algorithm converges monotonically.

*Proof.* For our ADDL problem in (13), variables $D$, $S$, $\Lambda$, $P$ and $W$ should be optimized. From Eqs.(14), (16), (18) and (20), we know that if $S$ and $W$ are fixed, the variables $\Lambda$, $D$ and $P$ can be optimized respectively, and $\Lambda$, $D$ or $P$ can be regarded as a single variable. On the contrary, if $\Lambda$, $D$ and $P$ are fixed, variables $S$ and $W$ can be optimized respectively, and similarly $S$ or $W$ can be treated as a single variable. So the optimization problem in (13) is a bi-convex problem over $\{(S,W),(\Lambda,P,D)\}$. According to [35], the optimal solutions of $(S,W)$ and $(\Lambda,P,D)$ correspond to the iteration steps in ACS, and the objective function (13) has a general lower bound 0. Thus, based on *Theorem 1*, the generated sequence $\{f(D^i, S^i, \Lambda^i, P^i, W^i)\}_{i \in t}$ by our ADDL algorithm is ensured to converge monotonically.

*Remark 2.* The sequence of $\{D^i, S^i, \Lambda^i, P^i, W^i\}_{i \in t}$ generated by our ADDL algorithm has at least one accumulation point. All the accumulation points are local optimal solutions of $f$ and have the same function value.

*Proof.* Suppose $\|S\|_F \to \infty$ or $\|P\|_F \to \infty$, then we can have $f(P,D,S,\Lambda,W) \to \infty$. Thus, $\{D^i, S^i, \Lambda^i, P^i, W^i\}_{i \in t}$ is bounded in a finite dimensional space, and then the compact set condition in *Theorem 2* (*Condition 1*) is satisfied. Thus, the sequence has at least one accumulation point. Moreover, for any $\tau > 0$, the optimizations of $S$ and $W$ in (15) and (19) are strictly convex and thus have unique solutions. So, based on *Theorem 2* (*Condition 2a*), all the accumulation points are local optimal and have the same function value.

*Remark 3.* Suppose $P$ and $D$ have unique solutions, then the sequence $\{D^i, S^i, P^i, W^i\}_{i \in t}$ generated by ADDL satisfies

$$\lim_{i \to \infty} \|D^{i+1} - D^i\| + \|S^{i+1} - S^i\| + \|\Lambda^{i+1} - \Lambda^i\| + \|P^{i+1} - P^i\| + \|W^{i+1} - W^i\| = 0 \quad (24)$$

*Proof.* Based on *Remark 2*, the *Conditions 1* and *2a* in the *Theorem 2* are satisfied in ADDL. If we have the unique optimal solution of $(\Lambda, P, D)$, then we have the conclusion (24) based on the *Condition 2b* in *Theorem 2* [35]. So it is easy to check that our ADDL is a reasonable approach.

*D. Classification Approach*

After the analysis projection $P = [P_1; \cdots P_l; \cdots P_c] \in \mathbb{R}^{K \times n}$ and analysis classifier $W = [W_1, \cdots W_l, \cdots W_c] \in \mathbb{R}^{c \times K}$ are obtained by our ADDL, outside new test data can be easily classified. Specifically, for a new test data $x_{new}$, we first compute its projective sparse codes by simply embedding it onto $P$, i.e., using $Px_{new}$ to approximate its sparse codes. Then, we further embed $Px_{new}$ onto the classifier $W$ in the form of $WPx_{new}$, so its soft label vector $f_{new}$ can be obtained as

$$f_{new} = WPx_{new} \in \mathbb{R}^{c \times 1}, \quad (25)$$

where the position of biggest entry in the soft label vector $f_{new}$ indicates the class labels of each test signal $x_{new}$. In other words, the *hard label* of each test signal $x_{new}$ can be assigned as $\arg\max_{i \leq c} (f_{new})_i$, where $(f_{new})_i$ represents the $i$-th entry of the soft label vector $f_{new}$.

## IV. DISCUSSION: RELATIONSHIP ANALYSIS

We show some important connections between our ADDL with other closely related DL algorithms.

*A. Connection to the DPL algorithm [3]*

Recall the objective function of our ADDL in (13), suppose the ideal condition that $P_l X_l = S_l$ is satisfied, if we further constrain $\alpha = 0$ and $\lambda = 0$, the problem in (13) becomes

$$\langle D, S, P \rangle = \arg\min_{D,S,P} \sum_{l=1}^{c} \left\{ \|X_l - D_l P_l X_l\|_F^2 + \tau \|P_l \overline{X_l}\|_F^2 + \tau \|S_l\|_{2,1} \right\}$$
$$s.t. \|d_v\|^2 = 1, v \in \{1,...,K\} \quad (26)$$

For (26), we use a $l_{2,1}$-norm constraint to regularize the coding coefficients so that sparse codes can be learnt. Note that if we ignore this sparse constraint, (26) is reduced to (2), which is just the objective function of DPL. Thus, DPL can be considered as a special case of our ADDL.

*B. Connection to the DLSI algorithm [15]*

By incorporating an incoherence promoting term into the category-specific dictionary learning approach, the existing DLSI method solves the following problem:

$$\langle D, S \rangle = \arg\min_{D,S} \sum_{l=1}^{c} \left\{ \|X_l - D_l S_l\|_F^2 + \upsilon \sum_{i}^{N_l} \|s_l^i\|_1 \right\} + \eta \sum_{l \neq t} \|D_l^T D_j\|_F^2, \quad (27)$$

where $\sum \|D_l^T D_j\|_F^2$ denotes the incoherence promoting term to encourage inter-class sub-dictionaries to be independent.

Note that our ADDL also ensure the sub-dictionaries of different classes to be independent as much as possible, but we use a different method. That is, we employ an analysis mechanism to minimize the reconstruction based on each sub-dictionary corresponding to class $l$ and the sparse codes of the training samples that do not belong to the class $l$, i.e., $\|D_l \overline{S_l}\|_F^2$, while DLSI aims to minimize $\sum \|D_l^T D_j\|_F^2$ directly.

Suppose the ideal conditions that the sub-dictionaries are independent, we can have $\sum \|D_l^T D_j\|_F^2 = 0$ and $\|D_l \overline{S_l}\|_F^2 = 0$. If we further constrain $P_l X_l = S_l$, $\|P_l \overline{X_l}\|_F^2 = 0$ and $\lambda = 0$ in our ADDL, our optimization problem can be transformed into

$$\langle D, S \rangle = \arg\min_{D,S,P} \sum_{l=1}^{c} \left\{ \|X_l - D_l S_l\|_F^2 + \tau \|S_l\|_{2,1} \right\}$$
$$s.t. \|d_v\|^2 = 1, v \in \{1,...,K\} \quad (28)$$

Clearly, we adopt a sparse $l_{2,1}$-norm constraint to ensure the learnt coding coefficients to be sparse. If we adopt the $l_1$-norm regularization, (28) is just the DLSI problem. Thus, DLSI is also considered as a special case of our ADDL.

*C. Connection to the LDL algorithm [12]*

Another related algorithm is *Latent Dictionary Learning* (LDL) [12] whose objective function is defined as

$$\langle D, S \rangle = \arg\min_{D,S} \sum_{l=1}^{c} \left\{ \|X_l - DT_l S_l\|_F^2 + \lambda_1 \|S_l\|_1 + \lambda_2 \|S_l - M_l\|_F^2 \right\}$$
$$+ \lambda_3 \sum_{l=1}^{c} \sum_{j \neq l} \sum_{p=1}^{K} \sum_{q \neq p} V_{l,q} (d_q d_p)^2 V_{j,p}, \quad (29)$$

where $T_l = diag(V_l)$, $V_l = [V_{l,1}, \cdots V_{l,K}]^T \in \mathbb{R}^{K \times 1}$ denotes a latent vector to indicate the relationship of all dictionary atoms to the $l$-th class data, $M_l$ is the mean matrix whose column vector is equal to the mean column vector of $S_l$. If all $V_l$ have only one non-zero value, each atom can only represent the data of a class, then the above problem can become

$$\langle D, S \rangle = \arg\min_{D,S} \sum_{l=1}^{c} \left\{ \|X_l - D_l S_l\|_F^2 + \lambda_1 \|S_l\|_1 + \lambda_2 \|S_l - M_l\|_F^2 \right\}$$
$$+ \lambda_3 \sum_{l=1}^{c} \sum_{j \neq l} \|D_l^T D_j\| \quad (30)$$

When $\lambda_2$ is constrained to 0, the above problem is just the DLSI problem [15]. So, LDL can also be considered as the special case of our ADDL algorithm.

*D. Connection to the JDL algorithm [36]*

Another related DL algorithm is called *Joint Dictionary Learning* (JDL) that learns a commonly shared dictionary and multiple class-specific dictionaries jointly. The criterion of the JDL algorithm is formulated as

$$\langle D_0, D_l, S_l \rangle = \arg\min_{D_0, D_l, S_l} \sum_{l=1}^{c} \left\{ \|X_l - [D_0, D_l] S_l\|_F^2 + \lambda_1 \|S_l\|_1 \right\} + \mu \Omega(S_1, \cdots S_c), \quad (31)$$

where $\Omega(S_1, \cdots S_c)$ is a discrimination promotion term based on the coding coefficients to encourage learnt coefficients to deliver small within-class scatter but large between-class scatter. Note that our ADDL also computes discriminative projective sparse codes by minimizing the analysis sparse codes extraction term. More specifically, we aim to use the sub-projection $P_l$ over class $l$ to embed the data from other classes $j$, $j \neq l$ to a nearly null space. Suppose the ideal condition that the original data are well represented by the learnt codes, then both the term $\Omega(S_1, \cdots S_c)$ in JDL and the codes extraction function in ADDL can be minimized to 0.

In addition, JDL considers the conditions that some of the samples from different classes may share certain common visual properties, so it jointly computes a shared dictionary to describe the common visual properties. Considering the ideal condition that the original signals are independent, the learnt shared dictionary by JDL is nearly null space. So, the criterion of JDL can be converted into

$$\langle D_l, S_l \rangle = \arg\min_{D_l, S_l} \sum_{l=1}^{c} \left\{ \|X_l - D_l S_l\|_F^2 + \lambda_1 \|S_l\|_1 \right\}. \quad (32)$$

Thus, the JDL can be considered as one special example of our presented ADDL formulation.

## V. SIMULATION RESULTS AND ANALYSIS

We mainly evaluate our proposed ADDL for representation and classification. The performance of our ADDL is mainly compared with several closely related SRC [4], DLSI [15], KSVD [7], D-KSVD [14], LC-KSVD [9], FDDL [11] and DPL [3]. Note that DLSI and KSVD did not define a clear classification method, thus we apply the same classification approach as SRC for DLSI and KSVD in this paper. In this study, nine groups of experiments are illustrated. The first group is to analyze the convergence of ADDL. The second one is to show some results about the parameter sensitivity analysis. The third, fourth, and fifth groups mainly exhibit the classification results on face, scene, and object datasets respectively. In the sixth and seventh part, we show some numerical results to compare the Mutual Coherence value and computational efficiency. We also explore the effect by using $l_{2,1}$-norm and $l_1$-norm regularizations. Note that the robustness analysis is evaluated in the last study.

We evaluate our ADDL on four real-world face databases: CMU PIE face database [37], MIT CBCL face recognition database [38], AR face database [39], UMIST face database [40], one object database: ETH80 database [42], and one scene category database: Fifteen Scene Categories database [41]. These datasets are widely used in existing works [1-3] [9][13-14] to evaluate DL methods. Details of these datasets are shown in Table II. For representation and classification on each database we split it into a training set and a testing set, where training set is formed by choosing fixed number of samples from each category, and treat the rest samples as test set. The training set is applied to learn the structured analysis dictionary and analysis classifier. The classifier is then used for evaluating the accuracies of the testing set. For fair comparison to the other existing DL methods, the accuracy of each method is averaged over 10 times different runs on the random splits of training and testing images to avoid the bias. We perform all the simulations on a PC with Intel (R) Core (TM) i3-4130 CPU @ 3.4 GHz 8G.

TABLE II.
DESCRIPTIONS OF USED REAL-WORLD IMAGE DATASETS.

| Dataset Name | # Samples | # Dim | # Classes |
|---|---|---|---|
| CMU PIE face | 11554 | 1024 | 68 |
| MIT CBCL face | 3240 | 1024 | 10 |
| UMIST face | 1012 | 1024 | 20 |
| AR face | 2600 | 540 | 100 |
| ETH80 object database | 3280 | 1024 | 80 |
| 15 Scene Categories | 4485 | 3000 | 15 |

### A. Convergence Analysis

In this experiment, we provide several numerical results about the convergence behavior of our ADDL algorithm. We mainly analyze the convergence behavior by describing the objective function values. All six datasets are employed. For the CMU PIE, MIT CBCL, UMIST face databases, and ETH80 object database, we randomly select 10 face images, 4 face images, 5 face images, and 6 object images from each subject for training respectively, and we simply set the number of dictionary as the number of training samples. For the AR database, we select 20 face image from each person class for training, and fix the number of dictionary atoms corresponding to an average of 5 items per person. For the fifteen scene category dataset, the training set consists of 1500 samples, i.e., 100 samples per class, and the number of dictionary consists of 450 items, i.e., 30 items from each individual. As for the other conditions, we adopt the same setting as Section V (*C*, *D* and *E*). All the results averaged over 30 times iterations are presented in Fig.5. Note that the iteration stops when the difference between two consecutive objective function values is less than 0.001.

From Fig. 5, we can find that the objective function value of our ADDL is non-increasing in the iterations, and more importantly it finally converges to a fixed value. Moreover, the number of iterations is often less than 20, which means that the convergence of our ADDL is relatively fast.

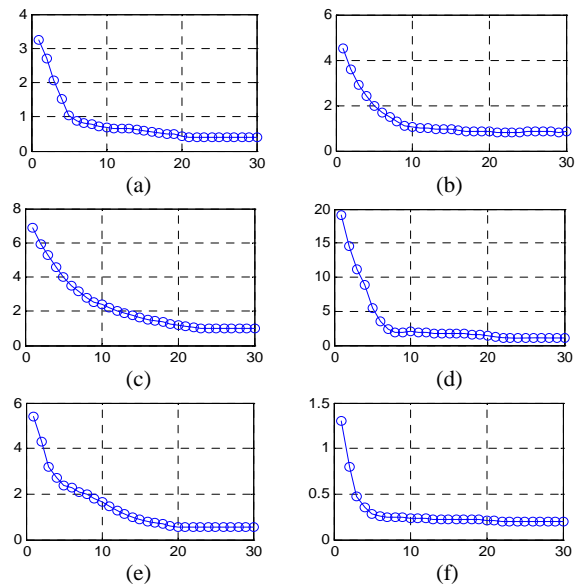

**Fig. 5:** Convergence behavior of our proposed ADDL method, where the *x*-axis represents the number of iterations, and the *y*-axis represents the objective function values. (a) The objective function value on CMU PIE; (b) The objective function value on MIT CBCL; (c) The objective function value on UMIST; (d) The objective function value on AR; (e) The objective function value on 15 scene categories; (f) The objective function value on ETH80 object database.

### B. Parameter Selection Analysis

Since our model has three parts, i.e., analysis incoherence promoting term, analysis sparse code extraction term and the analysis classifier training term, which are associated with the parameters $\tau, \alpha$ and $\lambda$, we would like to analyze the parameter sensitivity of our ADDL method in this study. Note that the parameter selection issue still remains an open problem. A heuristic way is select the most important ones. Specifically, we can fix one of the parameters and explore the effects of other two on the performance by grid search, and then change them within certain ranges if necessary.

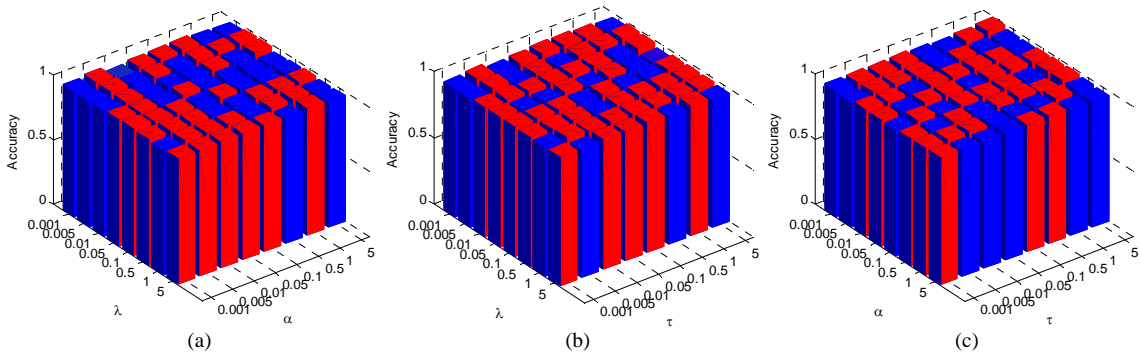

**Fig. 6:** Parameter sensitivity analysis of our ADDL on the MIT CBCL face database, where (a) the effects of tuning $\alpha$ and $\lambda$ on the recognition performance by fixing $\tau=0.05$; (b) the effects of tuning $\lambda$ and $\tau$ on the recognition performance by fixing $\alpha=0.1$; (c) the effects of tuning $\alpha$ and $\tau$ on the recognition performance by fixing $\lambda=0.001$.

In this study, the MIT CBCL face database is employed, and 4 face images of each individual are randomly chosen for training. The rests are used for testing. The number of dictionary atoms is equal to the number of training samples. For each pair of parameters, we average the results based on 10 randomly splits of training and testing data with varied parameters from $\{1\times10^{-3}, 5\times10^{-3}, 1\times10^{-2}, 5\times10^{-2}, 1\times10^{-1}, 5\times10^{-1}, 1, 5\}$. The parameter selection results are described in Fig.6, where three groups of results are presented. Specifically, the lower accuracies less than 97.5% are highlighted by blue color and higher accuracies more than 97.5% correspond to red color. We can find that our proposed ADDL performs well in a wide range of parameters selections in each group, which means ADDL is insensitive to the model parameters.

Also, we investigate the effects of the three terms on the results by setting $\tau=0$, $\alpha=0$ and $\lambda=0$ respectively. Note that if $\lambda=0$, there is no a clear classifier trained, so we apply the same classification method as DPL to classify new test data, i.e., when given a new test sample $x_{new}$, we identity the class label of $x_{new}$ by solving the following problem:

$$identity(x_{new}) = \arg\min_l \|x_{new} - D_l P_l x_{new}\|. \quad (33)$$

The results are shown in the following Table III. We can find clearly that when $\tau=0$ (i.e., the constraint $r(P_l, S_l)$ is removed), the learning performance of our algorithm is decreased significantly by delivering only 1% accuracy rate. This is because this term can bridge the reconstruction term ($\|X_l - D_l S_l\|_F^2$) with the analysis classifier training term, so removing this term means that the analysis linear classifier training term makes no sense if without the analysis sparse code extraction term. Suppose that the parameter $\alpha=0$ (i.e., the constraint $f(D_l)$ is removed) or $\lambda=0$ (i.e., the constraint $g(H_l, W_l, P_l)$ is removed), our method can achieve a better performance than the case of $\tau=0$, but is still inferior to the result of our ADDL with $\tau\neq0$, $\alpha\neq0$ and $\lambda\neq0$. Based on the above experimental analysis, we can conclude that the three terms or constraints are all important for improving the classification performance of our formulation. Specifically, the analysis sparse codes extraction part $r(P_l, S_l)$ associated with the parameter $\tau$ affects the results significantly.

Table III.
RECOGNITION RESULTS ON MIT CBCL FACE DATABASE.

| Compared Methods / Used Dataset | MIT CBCL |
|---|---|
| ADDL with $\alpha=0$, $\tau=0.05$, $\lambda=0.005$ | 97.6% |
| ADDL with $\alpha=1$, $\tau=0$, $\lambda=0.005$ | 1% |
| ADDL with $\alpha=1$, $\tau=0.05$, $\lambda=0$ | 97.4% |
| ADDL with $\alpha=1$, $\tau=0.05$, $\lambda=0.005$ | **98.3%** |

### C. Face Recognition

We test each method for face representation and recognition by dictionary learning. Four real face databases, i.e., CMU PIE, MIT CBCL, UMIST and AR, are evaluated. Some image examples of the face image databases are shown in Fig. 7. The face recognition results of our ADDL are mainly compared with those of SRC [4], DLSI [15], KSVD [7], D-KSVD [14], LC-KSVD [9], FDDL [11] and DPL [3]. For each compared method, we choose the parameters carefully. Since some dictionary learning algorithms (that is, KSVD, D-KSVD, and LC-KSVD) apply an $l_0$-norm based sparsity constraint, we also apply the constraint following their own for fair comparison.

### 1) Recognition results on CMU PIE database

In this section, we test each algorithm for face recognition using the CMU PIE face database. This database contains 68 persons with 41368 face images as a whole, and the face images were captured under varying pose, illumination and expression. Following the common evaluation procedure [1], 170 near frontal face images per person are employed for simulations. This subset consists of five near frontal poses (C05, C07, C09, C27 and C29) and all face images have different illuminations, lighting and expression.

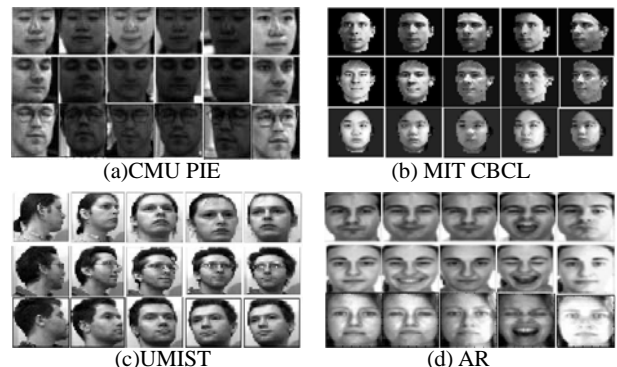

(a) CMU PIE  (b) MIT CBCL
(c) UMIST    (d) AR

**Fig. 7:** Examples of evaluated real face image databases.

TABLE IV.
RECOGNITION RESULTS USING PRINCIPAL FEATURES BY PCA ON THE CMU PIE FACE DATABASE.

| Number of training samples | 10 | 15 | 20 | 25 | 30 |
|---|---|---|---|---|---|
| SRC | 69.9 | 77.1 | 80.2 | 83.8 | 85.7 |
| DLSI | 60.7 | 73.3 | 80.1 | 82.3 | 83.4 |
| KSVD | 59.7 | 69.1 | 74.2 | 77.9 | 80.8 |
| D-KSVD | 61.3 | 72.1 | 79.4 | 84.4 | 85.8 |
| FDDL | 60.9 | 76.2 | 80.6 | 83.1 | 85.0 |
| LC-KSVD1 | 60.1 | 74.9 | 79.9 | 83.2 | 86.0 |
| LC-KSVD2 | 61.8 | 75.3 | 80.8 | 84.0 | 86.2 |
| DPL | 62.6 | 73.0 | 78.8 | 82.7 | 85.2 |
| Our ADDL | **81.7** | **88.5** | **90.0** | **91.1** | **91.8** |

TABLE V.
RECOGNITION RESULTS ON MIT CBCL FACE DATABASE.

| Evaluated Methods | 2 labeled Mean±STD(%) | 4 labeled Mean±STD(%) | 6 labeled Mean±STD(%) |
|---|---|---|---|
| SRC | 81.6±4.6 | 93.1±1.9 | 98.2±1.9 |
| DLSI | 83.6±4.9 | 94.5±2.0 | 98.6±1.3 |
| KSVD | 82.8±5.0 | 93.2±2.3 | 98.2±1.2 |
| D-KSVD | 83.9±5.2 | 95.8±3.6 | 98.7±0.9 |
| FDDL | 84.5±2.3 | 96.0±1.2 | 99.0±0.9 |
| LC-KSVD1 | 83.7±5.9 | 96.5±3.0 | 98.7±1.3 |
| LC-KSVD2 | 84.7±5.2 | 97.3±2.3 | 99.0±0.7 |
| DPL | 79.5±5.6 | 96.3±2.0 | 99.0±1.5 |
| Our ADDL | **86.2±3.4** | **98.3±1.2** | **99.3±0.9** |

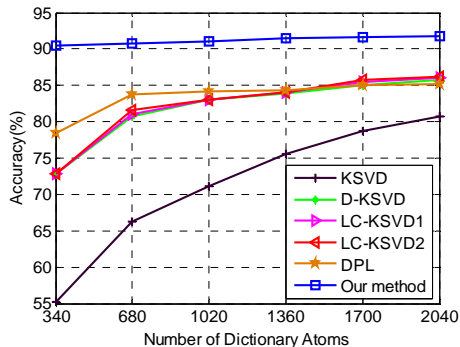

**Fig. 8.** Performance on CMU PIE with varying dictionary size.

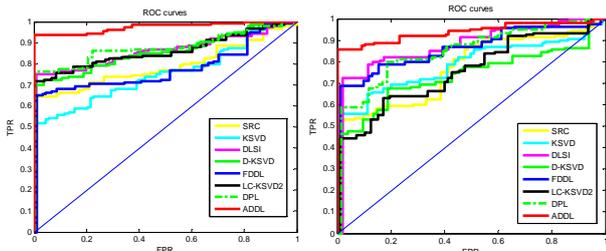

**Fig. 9.** ROC curves comparison of each method on CMU PIE: (left): the samples of first class are treated as the positive class data; (right): the samples of second class are treated as the positive class data.

Following [37], we choose the five near frontal poses (C05, C07, C09, C27, and C29) and use all images under different illuminations and facial expressions, and thus obtain 170 images for each person. As is common practice, all the face image are resized into 32×32 pixels. For the consideration of computational efficiency, we use *Principal Component Analysis* (PCA) [46] as a preprocessing step to reduce the number of dimensions by preserving 99% energy of the training data. For face recognition, we train on 5, 10, 15, 20, 25, and 30 images per person and test on the rest. To achieve the best performance, we simply set the number of dictionary atoms as its maximum (that is, the number of training images) in this study. We average the results based on 10 randomly splits of training and testing images in Table IV. We can find from the results in Table IV that our method is superior to all its competing methods in most cases, and more specifically outperforms the other by 4-5% improvement. It's also noted that our ADDL achieves 10% improvement than the others when the selected number of images for training is less than 20. Note that the main reason for the enhanced performance over the others is that the analysis mechanism adopted in ADDL ensures that the structured dictionary, projective sparse codes, and analysis classifier can be jointly gained. $\alpha=0.1, \tau=0.05$ and $\lambda=0.001$ are used in our proposed ADDL algorithm.

Additionally, we randomly choose 30 samples per class as training set and evaluate our ADDL with varying size $K$ of dictionary, i.e., $K$=340, 680, 1020, 1360, 1700, and 2040 in Fig.8 that indicates that our ADDL method maintains a higher recognition accuracy than its competitors even when the size of dictionary is relatively small.

To well show the comprehensive performance evaluation results of our ADDL with other methods, we also present the ROC curves [54] in this part. In this study, 10 samples are randomly selected from each class for training, and the rest ones are used for testing. The number of dictionary atoms is equal to the number of training samples. Note that ROC curves are mainly designed to evaluate the binary classification problem and CMU PIE face database contains 68 categories, thus in this study we mainly show two groups of results in Fig. 9. As can be seen, our ADDL is superior to other compared methods in most cases.

*2) Recognition results on MIT CBCL database*

In this study, the MIT CBCL face dataset is used for face recognition. This database provides two datasets: (1) High resolution pictures, i.e. frontal, half-profile and profile view; (2) Synthetic images (324 images per person, that is, 3240 face images totally) rendered from 3D head models. By following the common evaluation procedures, the second face set is tested. Note that we first normalize each sample to have unit $l_2$-norm for simulations. We randomly select 2, 4 and 6 images per person for training respectively, while testing on the rest. The number of atoms is simply set as its maximum for each method. We perform the tests over 10 randomly splits of training/testing images for each case, and we adopt the average of each run to be the final recognition rates. $\alpha=1, \tau=0.05$ and $\lambda=0.005$ are used in our model.

We illustrate the recognition results in Table V. We can observe that our ADDL achieves better performances than other methods in most cases. Specifically, when the training set is small, the superiority of our ADDL over the others is more obvious, i.e., achieving over 1.5% improvement when the training set contains 2 images per category.

Moreover, we compare the average recognition rates with varying the dictionary size $K$. The training set is formed by randomly choosing 6 images per person, while the testing set is the rest samples. $K$=20, 30, 40, 50 and 60 are utilized here. The experiment results are illustrated in Fig. 10. From the Fig. 10, we see that the average recognition rates of each method are increasing along with the increasing number of atoms. It can be found that ADDL outperforms its other competitors across all dictionary atoms.

We also present the ROC curves in this part. In this study, two samples are randomly selected from each class to form the training set, and test on the rest. The number of atoms is set to the number of training samples. Since the MIT CBCL face database contains 10 categories, so we mainly give two groups of results in Fig. 11 by regarding the samples of first class and samples of second class as the positive class data, respectively. As can be seen from the results, our proposed ADDL is still superior to its competitors.

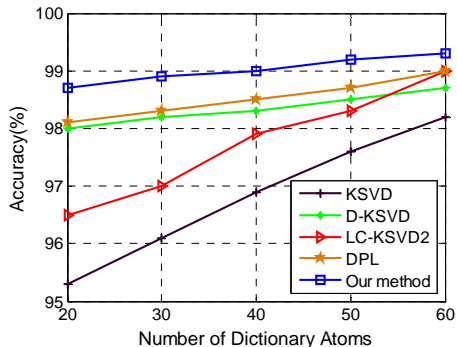

**Fig. 10**. Performance on MIT CBCL with varying dictionary sizes.

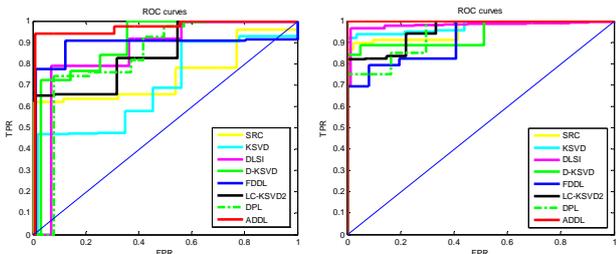

**Fig. 11**. ROC curves comparison of each method on MIT CBCL: (left): the samples of first class are treated as the positive class data; (right): the samples of second class are treated as the positive class data.

*3) Recognition results on UMIST database*

We evaluate each method for face recognition by utilizing the UMIST face database. UMIST face database contains 1012 images of 20 individuals. Each individual is shown in a range of poses from profile to frontal views and images are numbered consecutively as they were taken. Each image is resized into 32×32. Some examples are shown in Fig. 6c. In this experiment, we randomly select 6 images per class for training and the other images for testing. The number of dictionary atoms is equal to the number of training signals. We repeat the experiments 10 times with different randomly spits of the training and testing images to obtain reliable results. The final recognition rates are reported as the average of each run. In this simulation, Note that we first normalize each sample to have unit $l_2$-norm for simulations. $\alpha=0.01, \tau = 0.001$ and $\lambda = 0.001$ are used in our model.

The recognition results are shown in Table VI. We can find that our proposed ADDL achieves better performances than the competing methods. In addition, we also evaluate our ADDL by learning a smaller dictionary. It's noted that our method can outperform other competing methods when the dictionary consists of only 2 items per category.

TABLE VI.
RECOGNITION RESULTS ON UMIST FACE DATABASE.

| Evaluated Methods | Mean±STD(%) |
|---|---|
| SRC(5 items, 5 labeled) | 87.4±2.4 |
| DLSI(5 items, 5 labeled) | 87.1±2.1 |
| KSVD(5 items, 5 labeled) | 87.7±2.5 |
| DKSVD(5 items, 5 labeled) | 87.2±2.1 |
| FDDL(5 items, 5 labeled) | 87.5±1.6 |
| LC-KSVD1(5 items, 5 labeled) | 87.8±2.7 |
| LC-KSVD2(5 items, 5 labeled) | 88.6±2.0 |
| DPL(5 items, 5 labeled) | 88.9±1.6 |
| **ADDL(2 items, 5 labeled)** | **90.0±1.9** |
| **ADDL(5 items, 5 labeled)** | **90.9±1.7** |

TABLE VII.
RECOGNITION RESULTS USING RANDOM FACE FEATURES ON AR FACE DATABASE.

| Evaluated Methods | *Accuracy* |
|---|---|
| SRC (5 items, 20 labels) | 66.5% |
| KSVD(5 items, 20 labels) | 86.5% |
| DKSVD(5 items, 20 labels) | 88.8% |
| LLC(30 local bases, 20 labels)[16] | 69.5% |
| LLC(70 local bases, 20 labels)[16] | 88.7% |
| LC-KSVD1(5 items, 20 labels) | 92.5% |
| LC-KSVD2(5 items, 20 labels) | 93.7% |
| DLSI(5 items, 20 labels) | 93.1% |
| FDDL(5 items, 20 labels) | 95.6% |
| DPL(5 items, 20 labels) | 95.8% |
| JEDL(5 items, 20 labels) [1] | 96.2% |
| **ADDL(5 items, 20 labels)** | **97.0%** |

*4) Recognition results on AR database*

The AR face database is involved in this study. The AR face database contains over 4000 color images of 126 people. Each person has 26 face images taken during two sessions. Following the common evaluation procedure in [1-3][10], the face sample set including 2600 images of 50 males and 50 females is applied for simulations. Some examples are illustrated in Fig. 6d. Each face image has 165×120 pixels. In this study, we use the random face features [1-3][10], i.e., each face image is projected onto a 540-dimensional vector by a generated matrix from a zero-mean normal distribution, and each row of matrix is $l_2$ normalized. Following [1-3][10], we also randomly choose 20 images per person for training and the rest for testing. The dictionary contains 500 items, corresponding to an average of 5 items each category. $\alpha=0.01$, $\tau = 0.001$ and $\lambda = 0.001$ are set in our ADDL.

The recognition results are summarized in Table VII. Note that the results of the compared methods are adopted from [1-2]. We can find that our ADDL outperform other competing methods under the same experimental setting.

### D. Object Recognition on ETH80 database

In this section, we evaluate our ADDL on the ETH80 object database (totally 3280 images from 80 objects) [42]. This object dataset contains 8 big categories, including apple, car, cow, cup, dog, horse, pear, and tomato. In each big category, 10 subcategories are included, each of which contains 41 images from different viewpoints. Note that each image was resized to 32×32 pixels. Thus if each pixel in the images is treated as an input variable, each image is associated to a data point in a 1024-dimensional space. In this simulation, we perform dictionary learning over discriminant features, i.e., we use the *Linear Discriminant Analysis* (LDA) [43] to reduce the number of dimension. In the step of LDA, the generalized Eigen-problem is solved using SVD with the regularization parameter being 0.1, and 100 percent of the principal components are preserved in the PCA step, i.e., all non-zero eigenvalues are preserved. We randomly select 6 images per class for training and treat the rest as testing set. The number of dictionary atoms is set as its maximum for each method. We perform the tests over 10 randomly splits of training and testing images for each case, and employ the average of each run to be the final recognition accuracies. $\alpha=0.005$, $\tau = 0.001$ and $\lambda = 0.01$ are set in our model.

We describe the averaged classification results in Table VIII. We see that our method achieves better performances. It's also noticed that DPL also deliver a promising result that is comparable to our ADDL. Our method outperforms

TABLE VIII.
RECOGNITION RESULTS ON THE ETH80 OBJECT DATABASE.

| Evaluated Methods | Mean±STD(%) |
|---|---|
| SRC(6 items, 6 labeled) | 89.6±0.8 |
| DLSI(6 items, 6 labeled) | 92.7±0.9 |
| KSVD(6 items, 6 labeled) | 91.2±0.8 |
| DKSVD(6 items, 6 labeled) | 91.2±0.4 |
| FDDL(6 items, 6 labeled) | 93.2±0.3 |
| LC-KSVD1(6 items, 6 labeled) | 90.7±0.7 |
| LC-KSVD2(6 items, 6 labeled) | 91.5±0.8 |
| DPL(6 items, 6 labeled) | 97.7±0.2 |
| **ADDL(6 items, 6 labeled)** | **97.9±0.2** |

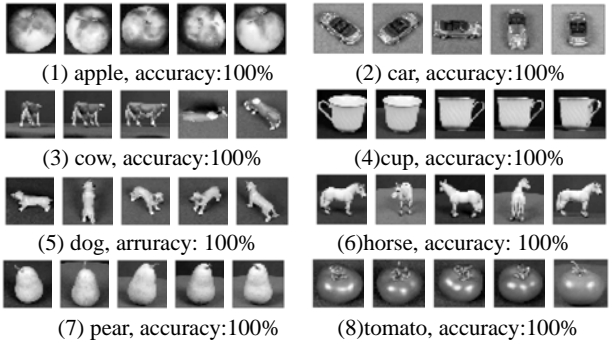

(1) apple, accuracy:100%   (2) car, accuracy:100%
(3) cow, accuracy:100%   (4) cup, accuracy:100%
(5) dog, arruracy: 100%   (6) horse, accuracy: 100%
(7) pear, accuracy:100%   (8) tomato, accuracy:100%

**Fig. 12:** Images examples from classes with high classification accuracy from the ETH80 object database.

other remains by more than 6% improvement. In addition, we also evaluate the recognition rates for individual classes, from which we see that there are several classes having 100 percent classification accuracy. Some instances of 8 classes with 100 percent of accuracies are shown in Fig. 12.

### E. Scene Category Recognition

The fifteen natural scene categories dataset [41] is involved in this section. This dataset includes 15 scenes, i.e., suburb, open country, mountain, coast, highway, forest, store, office, kitchen, industrial, living room, tall building, bedroom, street, inside city. Each category consists of 200 to 400 images, and each image has about 250×300 pixels. Similar to [2], the spatial pyramid feature using a four-level spatial pyramid and a SIFT-descriptor codebook with a size of 200 are obtained. The final spatial pyramid features are reduced to 3000 by PCA. Following the common settings [2][10], we randomly select 100 images per category for training and use the rest for testing. The dictionary size is 450 items, corresponding to an average of 30 items per category. In this study, $\alpha=0.1$, $\tau=0.05$ and $\lambda=0.001$ are used in ADDL.

Note that we directly adopt the results of the compared methods from [2]. The recognition results averaged over 10

Table IX.
RECOGNITION RESULTS USING SPATIAL PYRAMID FEATURES ON THE FIFTEEN SCENE CATEGORY DATABASE.

| Evaluate Methods | Accuracy |
|---|---|
| SRC(30 items, 100 labels) | 91.8% |
| KSVD(30 items, 100 labels) | 86.7% |
| DKSVD(30 items, 100 labels) | 89.1% |
| LC-KSVD1(30 items, 100 labels) | 90.4% |
| LC-KSVD2(30 items, 100 labels) | 92.9% |
| DLSI(30 items, 100 labels) | 92.5% |
| FDDL(30 items, 100 labels) | 93.1% |
| DPL(30 items, 100 labels) | 96.9% |
| **ADDL(30 items, 50 labels)** | **97.7%** |
| **ADDL(30 items, 100 labels)** | **98.1%** |

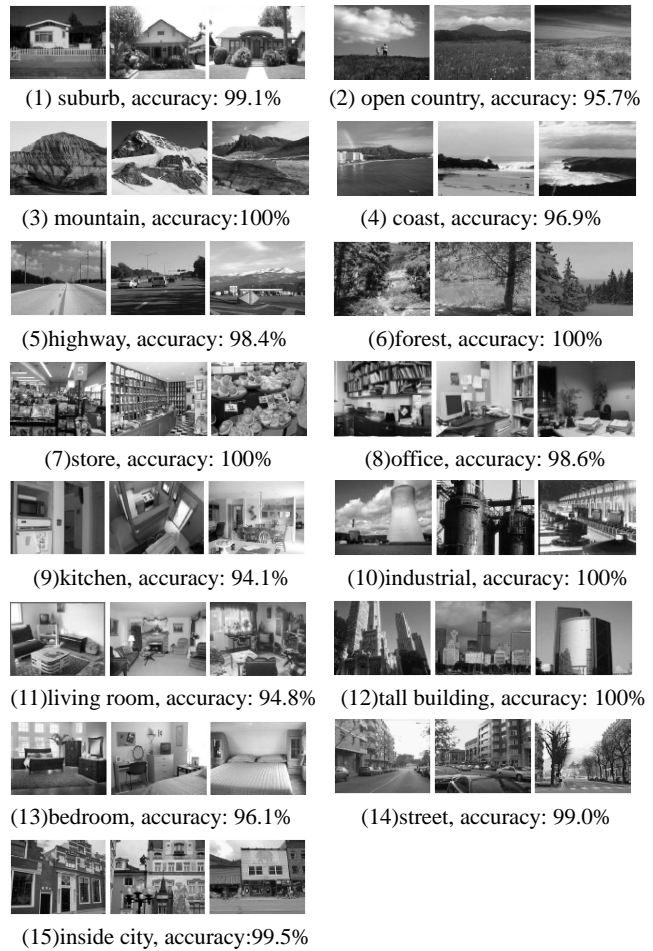

(1) suburb, accuracy: 99.1%   (2) open country, accuracy: 95.7%
(3) mountain, accuracy:100%   (4) coast, accuracy: 96.9%
(5) highway, accuracy: 98.4%   (6) forest, accuracy: 100%
(7) store, accuracy: 100%   (8) office, accuracy: 98.6%
(9) kitchen, accuracy: 94.1%   (10) industrial, accuracy: 100%
(11) living room, accuracy: 94.8%   (12) tall building, accuracy: 100%
(13) bedroom, accuracy: 96.1%   (14) street, accuracy: 99.0%
(15) inside city, accuracy:99.5%

**Fig. 13:** Image examples from classes from the fifteen natural scene categories database.

time repetitions are shown in Table IX. As seen from the Table IX, we can easily find that our ADDL obtains higher accuracies than the other models under the same setting. We also evaluate our ADDL using 50 images per class to train. In this case, our ADDL can also deliver better results than its competing methods. We also evaluate the recognition rates for individual classes. Fig.13 shows some examples with the accuracies of each individual. We can observe that most of the confusion occurs between the indoor classes, such as kitchen, bedroom, living room, and between some natural classes, for instance coast and open country.

### F. Comparison of Mutual Coherence value

We evaluate the mutual coherence values of our method and other competing methods in this study. Since we claim that the analytical incoherence promoting term can promote the learned inter-class sub-dictionaries to be independent as much as possible, we mainly compare the coherence values of inter-class sub-dictionaries. Specifically, we measure the coherence $\mu(D)$ of learned dictionary $D$ as the maximal correlation of any two atoms from various classes [47]:

$$\mu(D) = \max_{d_i \in D_i, d_j \in D_j, i \neq j} \left| \left\langle \frac{d_i}{\|d_i\|_2}, \frac{d_j}{\|d_j\|_2} \right\rangle \right|. \quad (34)$$

The value of $\mu$ function lies in [0, 1]. The minimum is reached for an orthogonal dictionary and the maximum for a dictionary containing at least two collinear atoms from different atoms. In this study, the fifteen scene categories dataset [41] is involved, and the same setting as Section *E* is adopted. The comparison results are illustrated in Fig. 14. As can be seen, DLSI, FDDL, DPL and ADDL algorithms have smaller coherence values than SRC, KSVD, DKSVD, and LC-KSVD2. Both DLSI and ADDL methods achieve the smallest coherence values, which means these two algorithms can learn the most independent sub-dictionaries, which is because both DLSI and our ADDL have taken the independence of sub-dictionary into account. But since our ADDL method can jointly learn a discriminative projection to extract coefficients and learn an analysis discriminative classifier for classification, it achieves a higher recognition rate than DLSI. It is noted that since SRC, KSVD, DKSVD, and LC-KSVD2 methods do not consider any incoherence promoting information, therefore these methods have higher coherence values. Since DKSVD jointly learn a multi-class classifier and LC-KSVD2 consider the discrimination of the coding coefficients further, they can deliver the satisfactory classification performances.

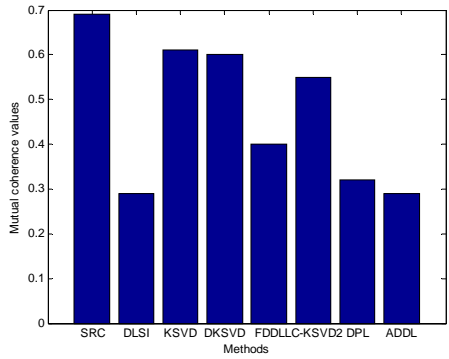

**Fig. 14:** Coherence comparison of each algorithm on the fifteen scene categories database.

### G. Comparison of Computational Time

We evaluate the training and testing time of our method and other competing methods in this experiment. We must note that the SRC algorithm does not learn any dictionary and utilize the original training data to perform classification, thus there is no training time for SRC. Considering that the same classification approach as SRC is used for KSVD, so we do not calculate the training and testing time of SRC for comparison. We mainly employ CMU PIE, AR, and ETH80 image databases for the experiments, and we use the same setting as *Section C* and *D*. We describe the computational time (training and testing time) of each method in Fig. 15.

From the Fig. 15a, we can find that our ADDL is more than 20 times faster than KSVD and D-KSVD, and about 100 times faster than LC-KSVD2 in the training phase. In testing, our ADDL is about 20 times faster than D-KSVD and LC-KSVD2, and is about 25 times faster than KSVD. From the Figs. 15b and 15c, we also find that KSVD, D-KSVD and LC-KSVD2 spend lots of time in both training and testing phase. The mean cause is that KSVD, D-KSVD and LC-KSVD adopted the costly $l_0$-norm constraint for computing the dictionary and sparse codes, and they have to involve an extra time-consuming sparse reconstruction process for each new test data. In particular, the training time of DPL is less than our ADDL. The main reason is that our ADDL jointly learns an analysis multi-class dictionary. As a result, our ADDL achieves better classification results by spending less computational time for both training and testing, compared with existing discriminative DL methods.

### H. Comparison of $l_{2,1}$-norm and $l_1$-norm Regularizations

Since we have claimed that our proposed ADDL adopted a sparse $l_{2,1}$-norm to regularize the coding coefficients for avoiding time-consuming $l_0$ or $l_1$-norm, we want to explore the effect of the $l_{2,1}$-norm and $l_1$-norm regularizations on the results of our method in this experiment. The CMU PIE and AR face databases are involved in this study. For CMU PIE, we randomly select 30 images per person for training and use the rest for testing. The size of dictionary is simply set as the number of training samples. For the AR database, we randomly select 20 images per person for training and use the rest for testing. The size of dictionary is set to 500. The comparison results of our ADDL method under $l_{2,1}$-norm and $l_1$-norm regularizations are shown in Table X. We can find that the classification accuracies of our ADDL under the two regularizations are similar. To well illustrate the superiority of the used $l_{2,1}$-norm, we also show the training and testing time in Fig.16. As can be seen, our ADDL under the $l_{2,1}$-norm regularization is much faster than the $l_1$-norm regularization under the case of comparable performance.

### I. Image Classification against Noisy Datasets

Note that our ADDL learns discriminative dictionary with the $l_{2,1}$-norm sparsity regularization rather than the robust $l_0$- and $l_1$-norm regularization, so we would like to test the robustness properties of our ADDL. Some numerical results are provided to illustrate the robustness properties of our method by comparing with several related DL methods for handling the images corrupted with noise. The MIT CBCL and UMIST face databases are involved in this study. For MIT CBCL database, we randomly select 6 face images per person for training and use the rest for testing. For UMIST database, we randomly select 5 face images per person for training and test on the rest. The size of dictionary is simply set as the number of training samples.

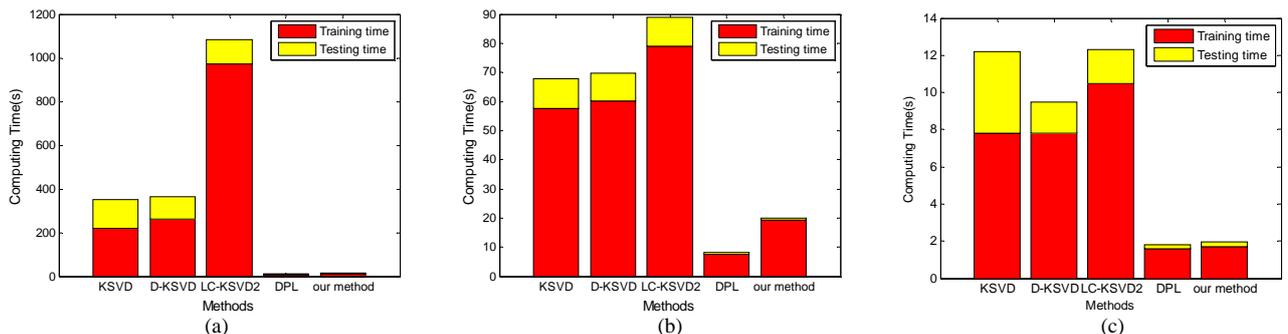

**Fig. 15:** Visualizations of needed time in training and testing phases. (a) CMU PIE face (training set: 2040 images, testing set: 9514 images); (b) AR face (training set: 2000 images, testing set: 60 images); (c) ETH80 object (training set: 480 images, testing set: 2800 images)

Table X.
RECOGNITION RESULTS USING $l_{2,1}$-NORM AND $l_1$-NORM.

| Regularized method | CMU PIE | AR |
|---|---|---|
| $l_{2,1}$-norm regularization | 91.8% | 97.0% |
| $l_1$-norm regularization | 91.5% | 97.1% |

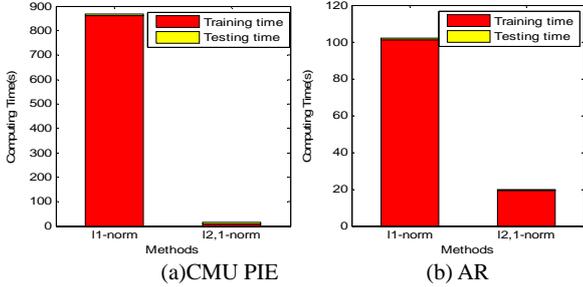

(a) CMU PIE  (b) AR

**Fig.16:** Visualizations of needed time in training and testing phases by $l_{2,1}$-norm and $l_1$-norm regularization.

Note that the random Gaussian noise is manually added into each image by using $Data = Data + \sqrt{Variance} \times randn(size(Data))$. We present the results under noisy cases by setting the value of *Variance*=200, 400, 600, 800, and 1000, respectively in Fig. 17. We can find that the classification accuracy of each method is decreased with the increasing values of *Variance*. It should be noted that our proposed ADDL delivers higher accuracies than the other methods in most cases. That is, our ADDL is more robust to the noise corruptions than the others because of the more reasonable formulation by analysis mechanism.

## VI. CONCLUDING REMARKS

We proposed an analysis mechanism based novel structured analysis discriminative dictionary learning approach that jointly integrates structured dictionary learning, projective sparse representation and analysis classifier training into a unified framework. To learn the structured dictionary, we incorporate an analytical incoherence promoting term to minimize the reconstruction error over each sub-dictionary and the sparse codes corresponding to different classes. We also consider to reduce the computational cost in training phase and testing phase when seeking the sparse coding coefficients. Specifically, we adopt the $l_{2,1}$-norm sparsity regularization rather than costly $l_0/l_1$-norm regularization and we train a classifier over approximated sparse codes via embedding instead of directly based on the original sparse codes. We also include an analysis classifier training term to enhance the discriminating power of the classifier.

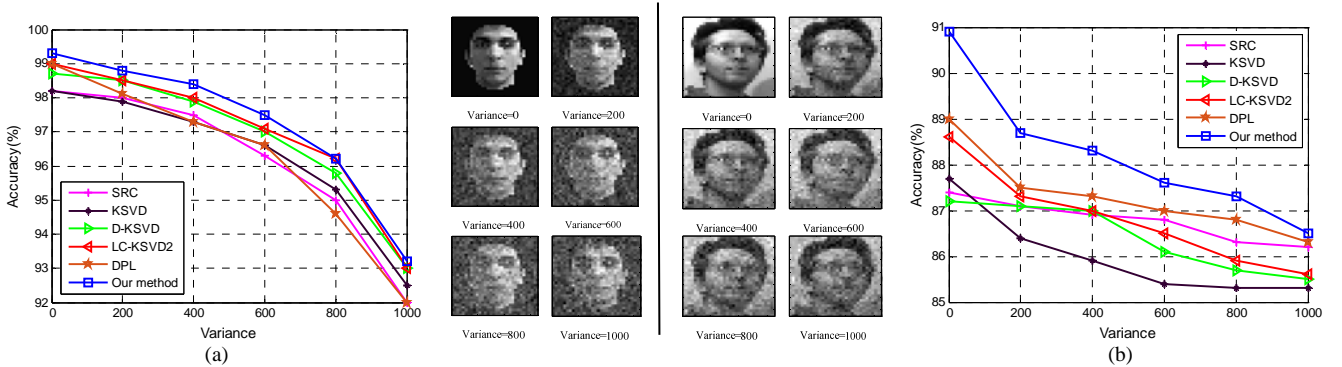

**Fig. 17:** Classification performance of each algorithm with varying *Variance* on two real image databases: (a) MIT CBCL (b) UMIST.

We mainly examined the effectiveness of our method on several widely-used real image databases. The experimental results demonstrate superior performances of our method in terms of performance and time efficiency, compared with several dictionary learning methods. But similar to existing DL methods, our method also performs dictionary learning based on the original data (e.g., real images) that usually contains noise, unfavorable features and even corruptions, which may have negative effects directly on the subsequent representation and classification performances. As a result, the representation and classification results may be reduced due to the negative effects. In future, we will discuss how to improve the robustness property against noise further. In addition, exploring how to extend our method to the semi-supervised scenario is crucial, since the number of labeled data is typically small in practice [7][44-45]. Extending our method to other application areas is also required.

## ACKNOWLEDGMENTS

The authors would like to express sincere thanks to the anonymous reviewers' comments that have made this paper a higher standard. This work is partially supported by the National Natural Science Foundation of China (61402310, 61672365, 61672364), Major Program of Natural Science Foundation of Jiangsu Higher Education Institutions of China (15KJA520002), Special Funding of Postdoctoral Science Foundation of China (2016T90494), Postdoctoral Science Foundation of China (2015M580462), Postdoctoral Science Foundation of Jiangsu Province (1501091B). Dr. Zhao Zhang is the corresponding author of this paper.

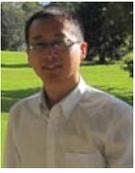
**Zhao Zhang** (M'13- ) received the Ph.D. degree from the Department of Electronic Engineering (EE), City University of Hong Kong, in 2013. He is currently an Associate Professor at the School of Computer Science and Technology, Soochow University, Suzhou, P. R. China. Dr. Zhang was a Visiting Research Engineer at the National University of Singapore, from Feb to May 2012. He then visited the National Laboratory of Pattern Recognition (NLPR) at Chinese Academy of Sciences (CAS), from Sep to Dec 2012. His current research interests include data mining & statistical machine learning, pattern recognition & image analysis. Dr. Zhang has authored/ co-authored over 60 technical papers published at prestigious international journals and conferences, including IEEE TKDE (3), IEEE TIP (3), IEEE TNNLS (2), ACM TIST, IEEE TSP, IEEE TCYB, IEEE TII (2), Pattern Recognition (4), Neural Networks (5), ICDM, ACM ICMR, ICIP and ICPR, etc. Specifically, he has got 11 first-author regular papers published by the IEEE/ACM Transactions journals. Dr. Zhang is now serving on the Editorial Board of Neurocomputing, IET Image Processing, Neural Processing Letters (NPL), and Neural Computing and Applications (NCA). Besides, he served as a Senior Program Committee (SPC) member of PAKDD 2017, an Area Chair (AC) of BMVC 2016/2015, a Program Committee (PC) member for several popular international conferences (including SDM 2017/2016/2015, PAKDD 2016, IJCNN 2017/2015, CAIP 2015, etc), and also often got invited as a journal reviewer for IEEE TNNLS、IEEE TIP、IEEE TKDE、IEEE TSP、IEEE TCYB、IEEE TMM, IEEE TII、Pattern Recognition、Information Sciences, etc.

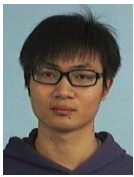
**Weiming Jiang** is now working toward the research degree at School of Computer Science and Technology, Soochow University, P. R. China. His current research interests mainly include machine learning, data mining and pattern recognition. He has authored or co-authored several papers published in the IEEE Transactions on Industrial Informatics (IEEE TII), IEEE Transactions on Neural Networks and Learning Systems (IEEE TNNLS), ACM International Conference on Multimedia Retrieval (ACM ICMR), IEEE International Conference on Data Mining (ICDM), and Pacific Rim Conference on Multimedia (PCM), etc.

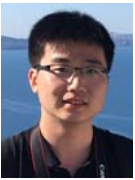
**Jie Qin** is currently a postdoctoral researcher with the Computer Vision Laboratory, ETH Zürich, Switzerland. He received the Ph.D. degree (under the supervision of Prof. Yunhong Wang) and B.E. degree from School of Computer Science and Engineering, Beihang University, China, in 2011 and 2017 respectively. From 2014 to 2015, he was a visiting researcher with the University of Sheffield, Supervised by Prof. Ling Shao, U.K. His research interests include Computer Vision, Machine Learning, Pattern Recognition, Image/Video Processing, and Human Activity Analysis. Dr. Qin has published papers in IEEE Conference on Computer Vision and Pattern Recognition (CVPR), IEEE Transactions on Image Processing (TIP), and Computer Vision and Image Understanding (CVIU), etc.

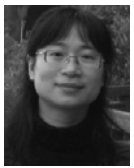
**Li Zhang** (M'08- ) is currently a Full Professor at the School of Computer Science and Technology, Soochow University, Suzhou, China. She obtained her Bachelor and PhD degrees from Xidian University in 1997 and 2002, respectively. She was also a postdoc researcher at the Shanghai Jiaotong University from 2003 to 2005. His research interests include pattern recognition, machine learning, and data mining. Dr. Zhang has authored/ co-authored more than 90 technical papers published at prestigious international journals and conferences, including IEEE TNNLS, IEEE TSP, IEEE Trans. SMC Part B, Information Sciences, Pattern Recognition, Neural Networks, etc.

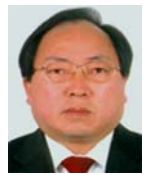
**Fanzhang Li** is now a full Professor and Dean of the School of Computer Science and Technology, Soochow University, P. R. China, and is also an adjunct professor at Beijing Jiaotong University, China. Prof. Li received his Master degree in engineering from the Department of Computer Science, The University of Science and Technology of China, in 1999. His current interests include Lie Group Machine learning, Data Mining, and Dynamic Fuzzy Logic. Prof. Li has published one Monograph Book on Dynamic Fuzzy Logic and its Applications, and has also published more than 100 technical papers in international journals and conferences.

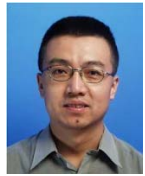
**Min Zhang** is now a Distinguished Professor and Director of the Research Institute of Intelligent Computing at Soochow University (China). He received his Bachelor degree and Ph.D. degree in computer science from Harbin Institute of Technology in 1991 and 1997, respectively. From 1997 to 1999, he worked as a postdoctoral research fellow in Korean Advanced Institute of Science and Technology in South Korea. He began his academic and industrial career as a researcher at Lernout & Hauspie Asia Pacific (Singapore) in September 1999. He joined Infotalk Technology (Singapore) as a researcher in 2001 and became a senior research manager in 2002. He joined the Institute for Infocomm Research (Singapore) as a research scientist in December 2003. His current research interests are machine translation, natural language processing, information extraction, intelligent computing and statistical machine learning. He has co-authored more than 130 papers in leading journals and conferences, and co-edited 10 books published by Springer and IEEE. He has been actively contributing to the research community as the conference chairs, program chairs and organizing chairs, and giving talks at conferences and lectures. He is the vice president of COLIPS (2011-2013), a steering committee member of PACLIC (2011~now), an executive member of AFNLP (2013~2014) and a member of ACL (2006~).

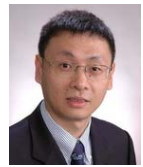
**Shuicheng Yan** (F'16- ) is currently a (Dean's Chair) Associate Professor in the Department of Electrical and Computer Engineering at National University of Singapore, and the founding lead of the Learning and Vision Research Group (http://www.lv-nus.org). Dr. Yan's research areas include computer vision, multimedia and machine learning, and he has authored /co-authored over 370 technical papers over a wide range of research topics, with Google Scholar citation >12,000 times and H-index-47. He is an associate editor (AE) of Journal of Computer Vision and Image Understanding, IEEE Transactions on Knowledge and Data Engineering (TKDE), IEEE Transactions on Circuits and Systems for Video Technology (IEEE TCSVT) and ACM Transactions on Intelligent Systems and Technology (ACM TIST), and has been serving as the guest editor of the special issues for TMM and CVIU. He received the Best Paper Awards from ACM MM13 (Best Paper and Best Student Paper), ACM MM'12 (demo), PCM'11, ACM MM'10, ICME'10 and ICIMCS'09, the winner prizes of the classification task in PASCAL VOC 2010-2012, the winner prize of the segmentation task in PASCAL VOC 2012, the honorable mention prize of the detection task in PASCAL VOC'10, 2010 TCSVT Best Associate Editor (BAE) Award, 2010 Young Faculty Research Award, 2011 Singapore Young Scientist Award, 2012 NUS Young Researcher Award, and the co-author of the best student paper awards of PREMIA'09, PREMIA'11 and PREMIA'12. He shall or has been General/Program Co-chair of PCM13, and ACM MM15, ICMR17 and ACM MM17.